\documentclass[10pt]{article}

\usepackage[margin=1in,top=1in,bottom=1in]{geometry}
\usepackage{amsmath,amssymb,amsthm}
\usepackage{graphicx}
\usepackage{booktabs}
\usepackage{hyperref}
\usepackage{xcolor}
\usepackage{algorithm}
\usepackage{algpseudocode}
\usepackage{subcaption}
\usepackage{enumitem}
\usepackage{cite}
\usepackage{url}
\usepackage{fancyhdr}
\usepackage{titlesec}
\usepackage[protrusion=true,expansion=false]{microtype}
\usepackage{parskip}

\hypersetup{
    colorlinks=true,
    linkcolor=blue!70!black,
    citecolor=green!50!black,
    urlcolor=blue!70!black
}

\titleformat{\section}{\large\bfseries}{\thesection}{1em}{}
\titleformat{\subsection}{\normalsize\bfseries}{\thesubsection}{1em}{}
\titleformat{\subsubsection}{\normalsize}{\thesubsubsection}{1em}{}

\algnewcommand{\algorithmicinput}{\textbf{Input:}}
\algnewcommand{\algorithmicoutput}{\textbf{Output:}}
\algnewcommand{\INPUT}{\item[\algorithmicinput]}
\algnewcommand{\OUTPUT}{\item[\algorithmicoutput]}

\title{\textbf{Home3D 1.0: A High-Fidelity Image-to-3D Asset\\Generation System for Interior Design}}

\author{
  Alibaba Group / Taobao\\
  \hyperref[app:contributions]{\texttt{\{Home3D-team\}@alibaba-inc.com}}
}

\date{}

\begin{document}

\maketitle
\vspace{-1.5em}

\begin{abstract}
We present \textbf{Home3D 1.0}, a modular image-to-3D generation system that produces
high-quality 3D assets from a single reference image, targeting interior design and
e-commerce applications. Given a photograph of a
furniture or decor item, the system outputs a mesh with physically-based rendering (PBR)
materials, and the mesh can be decomposed into material-specific components. The pipeline is organized into four tightly coupled modules:
\emph{Geometry} reconstructs a watertight mesh through latent SDF modelling with a
geometry VAE and a coarse-to-fine flow-matching DiT;
\emph{Texture} predicts multiview albedo observations, reprojects them onto the mesh,
and completes unseen surface regions with a 3D texture field;
\emph{Material} uses MatWeaver to obtain component masks through video-based
segmentation and UV-space voting, then retrieves and bakes PBR maps from a curated
material library through hierarchical multi-modal matching; and
\emph{Parts} generates material-editable semantic part meshes with a PartVAE and
PartDiT, decoding multi-head part-specific SDF fields in one pass.
Each module is evaluated independently with dedicated metrics, highlighting both
the current system capability and the remaining gaps toward broader deployment.
\end{abstract}

\begin{center}
\href{https://homeai.taobao.com}{\texttt{Home3D}}
\end{center}

\begin{figure}[H]
  \centering
  \includegraphics[width=\textwidth,height=0.4\textheight,keepaspectratio]{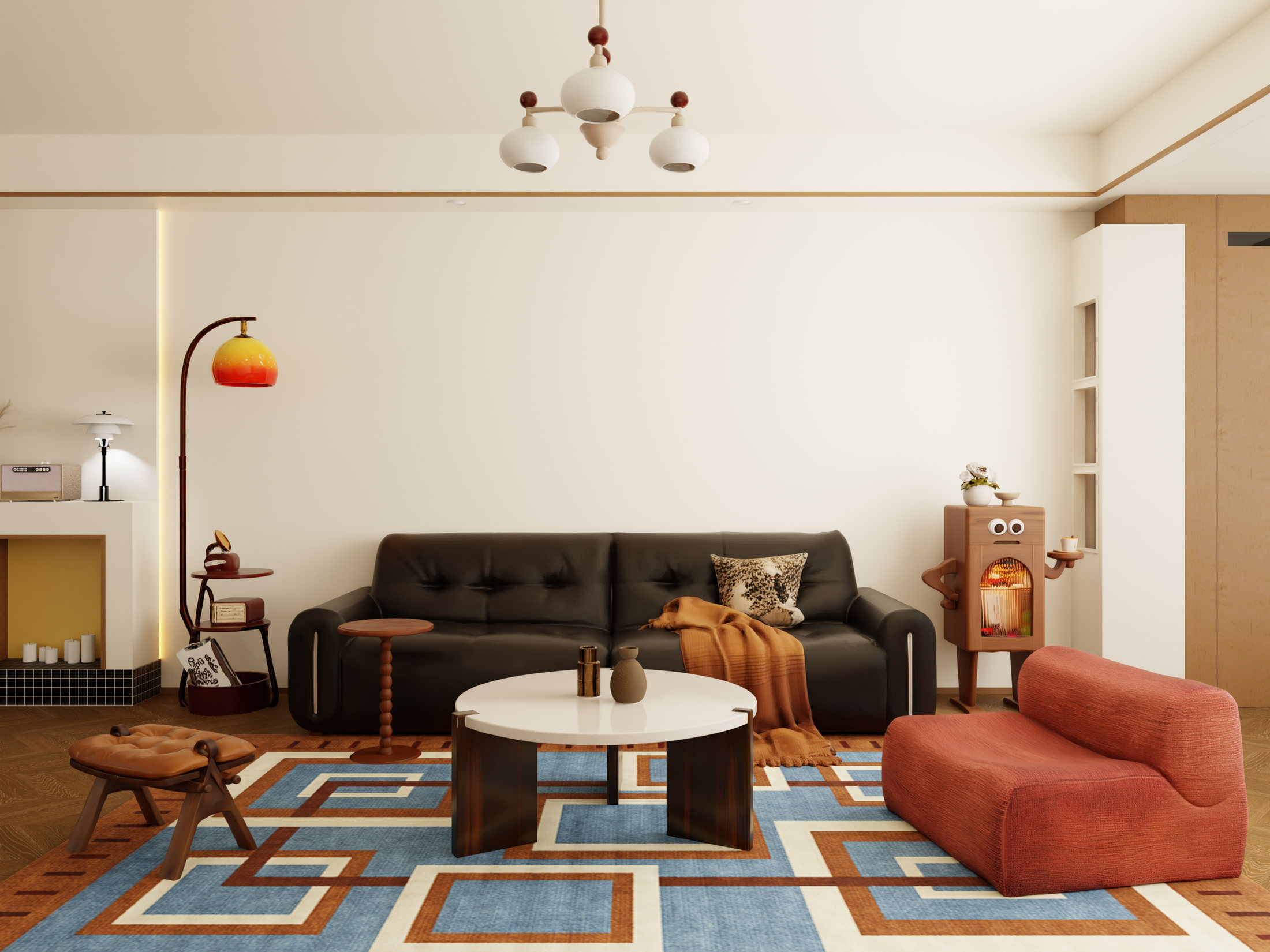}
  \caption{A living room scene populated with furniture models generated by Home3D 1.0.}
  \label{fig:gallery}
\end{figure}

\enlargethispage{2\baselineskip}
\newpage
\tableofcontents
\clearpage

%% ====================================================================
\section{Introduction}
\label{sec:intro}
%% ====================================================================

Interior design turns image-to-3D generation into a product modeling problem. In
generic 3D generation, it is often acceptable to produce a plausible object that
belongs to the requested category~\cite{seed3d1,hunyuan3d21,hunyuan3dStudio2025,tripoSR2024}. For
furniture and home decoration, this is not enough. The generated asset is expected to
represent a specific product that may be used in room design, e-commerce display, handcrafted editing and
material replacement. A generated chair should not merely look like a chair, it should
preserve the reference chair's silhouette, proportions, style, construction logic, and
visible finish.

Furniture is therefore different from arbitrary 3D object categories. It has stable
real-world scale, canonical orientation, floor or wall contact constraints, repeated
rigid structures, large planar or smoothly upholstered surfaces, and category-specific
details such as legs, cushions, handles, shelves, seams, bevels, and thin supporting
frames. These properties make furniture especially sensitive to common 3D generation
errors. A slightly noisy tabletop, broken chair leg, distorted cabinet panel, missing
handle, or over-smoothed cushion can immediately make the asset unusable for interior
design.

Current 3D generation methods remain insufficient for this setting. Mesh quality is
often unstable: generated furniture may contain noisy surfaces, missing structures,
weak symmetry, broken thin parts, or over-smoothed geometric details. Texture quality
is also limited: fabric weave, leather grain, wood texture, stone veins, seams,
stitching, and decorative patterns are frequently blurred or inconsistent across
views. Material quality is even harder to guarantee, because directly generated PBR
maps often fail to recover accurate albedo, roughness, metallic response, normal
detail, and true material identity. Finally, existing part generation methods usually
decompose furniture by function, such as legs, backs, seats, drawers, and handles.
It is not the enough abstraction
for interior design. Designers usually edit furniture by material: fabric, wood,
metal, glass, leather, stone, or paint. A material-aware decomposition is therefore
more useful than a purely function-aware one.

We present \textbf{Home3D 1.0}, a modular image-to-3D generation system purpose-built
for interior furniture and decoration assets. Home3D 1.0 takes a single reference image as
input and produces a high-quality 3D asset with watertight geometry, full texture
coverage, material-aware regions, retrieved PBR materials, and material-aware part
structure. The central idea is simple: furniture generation should preserve product
identity while making materials editable. To achieve this, Home3D 1.0 separates the
problem into specialized modules instead of relying on a single monolithic model:

\begin{itemize}
\item \textbf{Product-Faithful Geometry and Texture Generation.} We achieve this with a coarse-to-fine latent SDF
generation pipeline and a multiview texture pipeline. The geometry branch first
establishes a globally coherent furniture shape and then refines sharp boundaries,
thin supports, and local surface details, while the texture branch predicts
multiview albedo, reprojects it onto the mesh, and completes occluded regions in 3D
surface space. On our furniture benchmark (which is derived from the same source of
3D-FRONT~\cite{threedfront}), Home3D 1.0 achieves the best geometry scores among the
evaluated image-to-3D systems, with CD $0.4936$ ($\times 10^{-3}$), EMD $5.174$
($\times 10^{-2}$), and F1@0.01 $0.6329$.

\item \textbf{High quality PBR Material Creation.} We introduce a material branch that segments the reconstructed surface
into material regions, retrieves high-quality PBR materials through hierarchical
visual-semantic search, and bakes metallic, roughness, normal, and other attributes
into compact engine-ready texture maps. This branch supports automatic large-scale
e-commerce asset creation from product photos while avoiding the instability of
generating every PBR map from scratch.

\item \textbf{Material-Aware Part Generation for Interior Editing.} Designers edit
furniture by material regions such as fabric, wood, metal, glass, leather, stone, and
paint, rather than only by functional parts. We therefore introduce a part generation
branch that produces semantic, material-editable part meshes with a multi-head SDF
representation. Compared with sequential part generation, the decoder directly
predicts part-specific SDF fields in one pass rather than running a separate generator
for each part, enabling designers to replace
component-level materials and refine furniture appearance when composing interior
scenes.
\end{itemize}

Leveraging this system, we can enable scalable production of 3D assets tailored to e-commerce 
applications, while simultaneously offering sophisticated editing capabilities to support 
interior design use cases.

%% ====================================================================
\section{Method}
\label{sec:method}
%% ====================================================================

\subsection{Overview}
\label{sec:method_overview}

\begin{figure}[h]
  \centering
  \includegraphics[width=\textwidth]{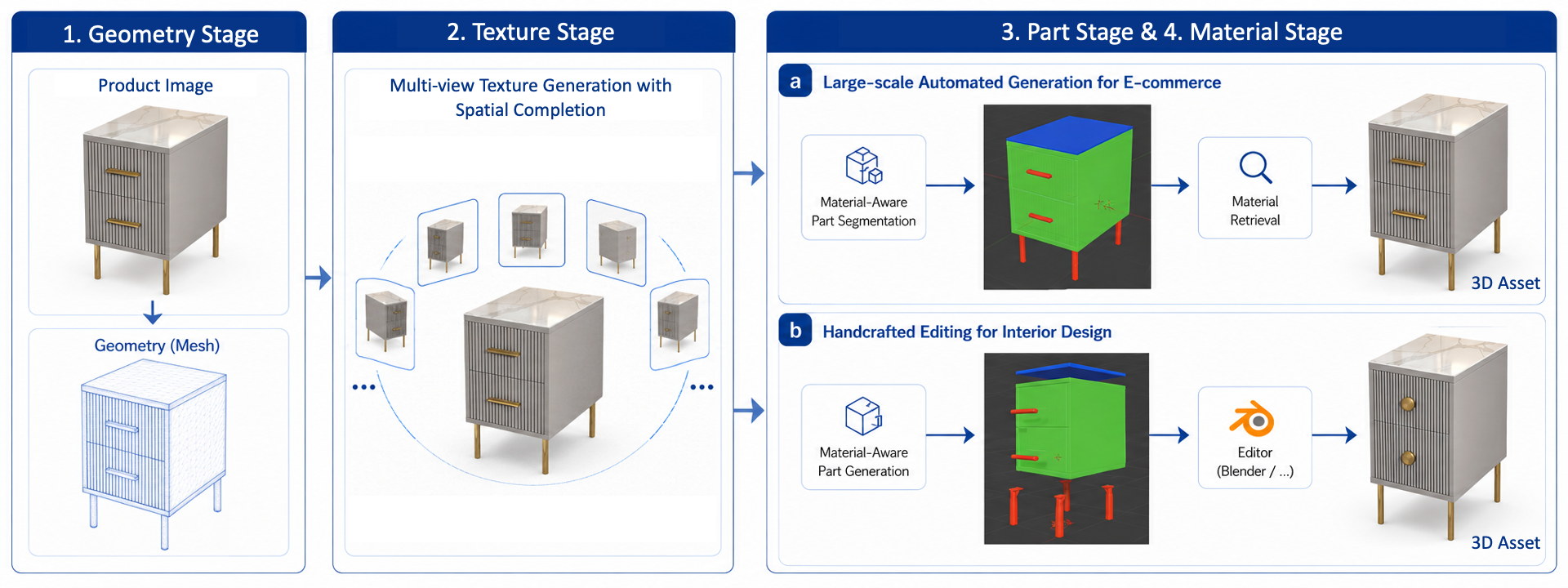}
  \caption{Home3D 1.0 System}
  \label{fig:pipeline}
\end{figure}

Home3D 1.0 decomposes furniture image-to-3D generation into four specialized stages:
geometry, texture, material, and parts. Given a single RGB reference image $I$, the
system reconstructs the product shape, recovers complete appearance, assigns
retrieved PBR materials to editable material regions, and produces perfectly intact part
structure. The overall flow is shown in Figure~\ref{fig:pipeline}.

The pipeline proceeds as follows:
\begin{enumerate}[leftmargin=*,topsep=2pt,itemsep=2pt]
  \item \textbf{Geometry} (\S\ref{sec:geometry}) reconstructs a watertight mesh
        $\mathcal{M}$ from the input image using a coarse-to-fine latent SDF
        generation process.
  \item \textbf{Texture} (\S\ref{sec:texture}) predicts multiview
        albedo observations, reprojects them onto $\mathcal{M}$, and completes
        unseen regions in 3D surface space.
  \item \textbf{Material} (\S\ref{sec:material}) segments material regions on the
        mesh and retrieves high-quality PBR maps from a curated material library
        instead of generating all material channels from scratch.
  \item \textbf{Parts} (\S\ref{sec:parts}) generates semantic part meshes
        conditioned on the reference image and reconstructed geometry, enabling
        material-level editing.
\end{enumerate}

Geometry, Texture, and Material run sequentially because each stage depends on the
previous output. Parts only requires the reference image and reconstructed geometry,
so it can run in parallel after Geometry. This modular design keeps each component
focused on one furniture-specific failure mode while assembling the outputs into one
coherent, editable asset.

\subsection{Geometry}
\label{sec:geometry}

\begin{figure*}[h]
    \centering
    \includegraphics[width=\textwidth]{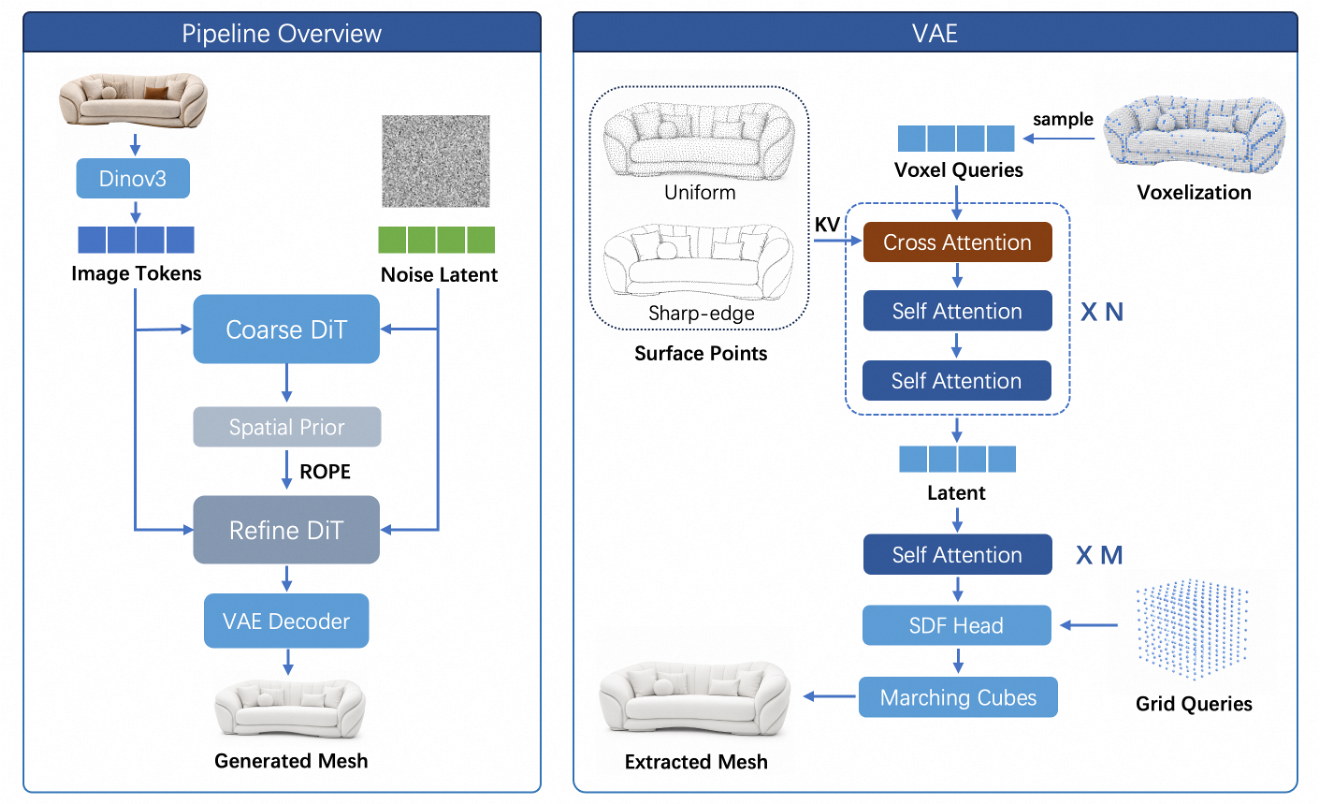}
    \caption{Overview of the geometry generation framework, including image-conditioned coarse-to-fine DiT latent generation and a geometry VAE that encodes surface samples and decodes SDF queries into meshes.}
    \label{fig:geo_framework}
\end{figure*}

As illustrated in Fig.~\ref{fig:geo_framework}, the geometry generation stage synthesizes high-fidelity 3D mesh geometry from a single RGB image 
by combining VAE-based latent geometry modeling with image-conditioned diffusion generation. 
Inspired by recent 3D generation methods such as LATTICE~\cite{lattice} and Seed3d 2.0~\cite{seed3d2}, 
the diffusion generation pipeline is formulated as a two-stage coarse-to-fine process. 
The first-stage \emph{Coarse DiT} focuses on producing a globally coherent coarse shape that 
preserves the object’s overall topology, global proportions, and semantic structure. 
The second-stage \emph{Refine DiT} takes the coarse geometry as a spatial prior 
and performs detail-oriented refinement, enabling the recovery of sharp boundaries, 
local geometric variations, and high-frequency surface structures. 

\subsubsection{Data}

The training data consists of open-source 3D models from ObjaverseXL~\cite{objaverseXL} 
and private 3D models, covering diverse categories such as furniture, characters, 
creatures, vehicles, architecture, props, and game assets. Compared with public datasets, 
the private data mainly consists of high-quality, production-grade furniture models sourced from
Tmall Design Home suppliers.

The dataset is divided into a large-scale pretraining set and a high-quality fine-tuning set. The pretraining set emphasizes broad distributional coverage, 
enabling the model to learn generalizable shape priors across categories. 
The fine-tuning set focuses on geometric cleanliness, rich surface details, 
and strong image–geometry consistency, thereby improving the final generation quality.

\subsubsection{Data Processing}

The data preprocessing pipeline consists of six stages: format canonicalization, quality filtering, 
geometric standardization, watertight reconstruction, sampling for supervision, and condition rendering.

First, to standardize asset formats, all raw assets are converted into a unified mesh representation. 
During this stage, empty nodes, duplicate faces, degenerate triangles, isolated components, pedestals, 
and irrelevant environmental structures are removed. For quality filtering, 
statistical features such as mesh connectivity, boundary completeness, scale distribution, 
and geometric complexity are used to eliminate simple primitives, noisy scanned assets, fragmented meshes, 
and obviously incomplete models.

During geometric standardization, each target model is first normalized to a unified 3D 
coordinate system within the range of $[-1,1]^3$. In addition, a lightweight deep network is trained using 
manually annotated front-facing orientation labels to predict the canonical orientation of each model. 
This allows all training samples to be aligned into a consistent canonical pose, 
reducing distributional shifts caused by pose variations and improving the stability of DiT-based shape prior learning.

In the watertight reconstruction stage, a GPU-accelerated high-resolution sparse SDF computation pipeline is adopted. 
A watertight mesh is then extracted using the MC algorithm. This process can repair open surfaces, cracks, 
self-intersections, thin shells, and part of the internal structures. 
A subsequent post-processing step further removes internal components that are weakly correlated with the visible outer surface, 
ensuring that the SDF supervision has a well-defined inside–outside semantic meaning.

For training sample construction, surface points, sharp-edge high-curvature points, near-surface points, 
and free-space points are sampled for each mesh. 
Specifically, surface points are sampled per object as input to the VAE encoder, 
including uniformly sampled surface points and sharp-edge high-curvature points. 
In addition, near-surface and free-space supervision points are sampled to compute the TSDF reconstruction loss. 
For condition image generation, Blender Cycles is used to render multi-view images as visual conditioning signals, with 15 views rendered for each object.

\subsubsection{Model Architecture}

\paragraph{VAE.}
As illustrated in Fig.~\ref{fig:geo_framework}, the geometry VAE learns a compact latent representation for continuous 3D geometry. 
Given a mesh, we sample surface points from both uniformly distributed regions and sharp-edge regions, producing an augmented point cloud 
$\mathbf{P}\in\mathbb{R}^{N\times 7}$, where each point contains its 3D position, surface normal, and a sharp-edge indicator. 
The encoder adopts a Perceiver-style latent query mechanism, in which a set of learnable anchor queries interact with point-cloud features through multi-layer cross-attention:
\begin{equation}
    \mathbf{Z}
    =
    \mathrm{Enc}_{\phi}(\mathbf{P})
    =
    \mathrm{CrossAttn}
    \left(
    \mathbf{Q}_{\mathrm{anchor}},
    \mathbf{K}_{\mathbf{P}},
    \mathbf{V}_{\mathbf{P}}
    \right)
    \in \mathbb{R}^{M\times d}.
    \label{eq:vae_encoder}
\end{equation}
Unlike shallow point aggregation in previous methods~\cite{lattice}, the latent tokens repeatedly attend to local geometric features, normal features, and edge-aware features across multiple layers. 
This design enables the VecSet latent representation $\mathbf{Z}$ to capture both global topology and high-frequency local structures, improving the reconstruction upper bound for thin parts, sharp boundaries, and local curvature variations. 
The decoder represents geometry as an implicit signed distance field conditioned on the latent tokens:
\begin{equation}
    s(\mathbf{x})
    =
    f_{\theta}(\mathbf{x}, \mathbf{Z}),
    \qquad
    \mathbf{x}\in\mathbb{R}^{3},
    \label{eq:sdf_decoder}
\end{equation}
where $s(\mathbf{x})$ denotes the predicted SDF value at query position $\mathbf{x}$. 
The final mesh is extracted from the zero-level set using Marching Cubes~\cite{marchingcubes1987}.

\paragraph{Flow Matching DiT.}
The image-conditioned generation model follows a coarse-to-fine Flow Matching DiT architecture. 
Given an input RGB image $\mathbf{I}$, a pretrained visual encoder extracts image tokens
\begin{equation}
    \mathbf{C}_{I} = E_{\mathrm{img}}(\mathbf{I}),
    \label{eq:image_tokens}
\end{equation}
which serve as the semantic condition for latent geometry generation. 
The Coarse DiT first generates a globally consistent VecSet latent $\widehat{\mathbf{Z}}_{c}$ from noise, focusing on object scale, category-level structure, and major topology:
\begin{equation}
    \widehat{\mathbf{Z}}_{c}
    =
    G_{\mathrm{coarse}}
    \left(
    \mathbf{Z}_{T}, \mathbf{C}_{I}
    \right),
    \qquad
    \mathbf{Z}_{T}\sim\mathcal{N}(\mathbf{0},\mathbf{I}).
    \label{eq:coarse_dit}
\end{equation}
The coarse latent is decoded into a preliminary mesh, which is then voxelized to obtain a spatial geometric prior. 
The voxelized prior is encoded with RoPE-based positional encoding:
\begin{equation}
    \mathbf{S}_{c}
    =
    \mathrm{RoPE}
    \left(
    \mathrm{Voxelize}
    \left(
    \mathrm{Dec}_{\theta}(\widehat{\mathbf{Z}}_{c})
    \right)
    \right).
    \label{eq:spatial_prior}
\end{equation}
Conditioned on the image tokens and the coarse spatial prior, the Refine DiT predicts a refined latent representation:
\begin{equation}
    \widehat{\mathbf{Z}}_{r}
    =
    G_{\mathrm{refine}}
    \left(
    \mathbf{Z}_{T}, \mathbf{C}_{I}, \mathbf{S}_{c}
    \right).
    \qquad
    \mathbf{Z}_{T}\sim\mathcal{N}(\mathbf{0},\mathbf{I}).
    \label{eq:refine_dit}
\end{equation}

This two-stage formulation separates global shape formation from local detail synthesis, thereby reducing the burden on a single generator and improving the fidelity of the final reconstructed mesh.

\subsubsection{Training Method}

The VAE is trained on the high-quality fine-tuning set to learn SDF reconstruction. The loss function includes SDF regression loss and KL-divergence loss:
\begin{equation}
  \mathcal{L}_{\mathrm{recon}} =
    \mathbb{E}\bigl[\mathrm{MSE}(s(\mathbf{x}),\, GT)\bigr]
    + \lambda_{\mathrm{KL}}\,\mathcal{L}_{\mathrm{KL}}
  \label{eq:geo_recon}
\end{equation}
with KL weight $\lambda_{\mathrm{KL}}=10^{-3}$ and 10k-step linear warmup.

Both DiT stages are trained with the flow matching objective~\cite{lipman2022flow}:
\begin{equation}
    \mathcal{L}_{\mathrm{FM}}
    =
    \mathbb{E}_{t,\mathbf{Z}_{0},\mathbf{Z}_{1}}
    \left[
    \left\|
    v_{\psi}(\mathbf{Z}_{t},t,\mathbf{C})
    -
    (\mathbf{Z}_{1}-\mathbf{Z}_{0})
    \right\|_{2}^{2}
    \right],
    \qquad
    \mathbf{Z}_{t}=(1-t)\mathbf{Z}_{0}+t\mathbf{Z}_{1},
    \label{eq:flow_matching}
\end{equation}
where $\mathbf{Z}_{0}$ is sampled from a Gaussian prior, $\mathbf{Z}_{1}$ is the target VAE latent, and $\mathbf{C}$ denotes the corresponding conditioning signals. 
Both DiT follows a progressive training schedule from large-scale pretraining to high-quality fine-tuning, 
gradually increasing both the number of latent tokens and the image resolution.

\subsection{Texture}
\label{sec:texture}
\subsubsection{Texture Generation Framework}
\label{sec:albedo_and_completion}

\begin{figure*}[h]
    \centering
    \includegraphics[width=\textwidth]{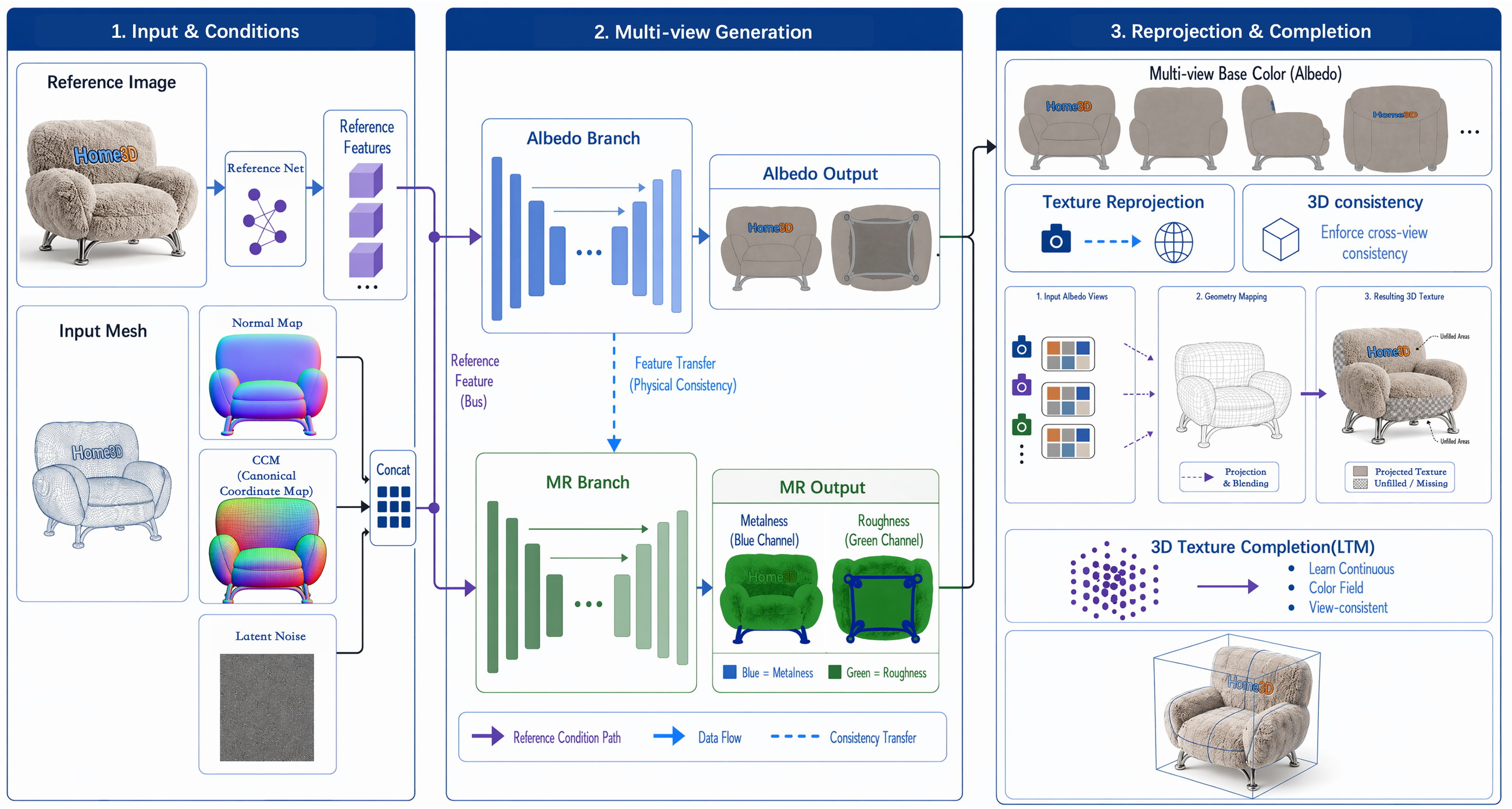}
    \caption{Overview of the proposed texture generation framework. The system predicts multiview albedo from a reference image, reprojects it onto the mesh surface, and completes missing regions to obtain a fully textured mesh.}
    \label{fig:framework}
\end{figure*}

As illustrated in Fig.~\ref{fig:framework}, we propose a unified pipeline for generating complete 3D textures from a single reference image. Given a reference image, the corresponding mesh, and its geometry images, the system first predicts albedo maps from multiple canonical viewpoints conditioned on the reference image and geometric information. Instead of generating each view independently, the model jointly predicts all target views, which helps preserve the appearance characteristics of the input, improve cross-view consistency, and reduce the influence of illumination effects such as shadows and highlights. The generation stage may also produce additional material-related outputs in parallel, but these auxiliary predictions are not involved in the texture completion process described in this section.

The predicted multiview albedo maps are then reprojected onto the target mesh using the geometric correspondence defined by the canonical views. This step transfers the image-space predictions into a unified 3D surface representation and produces a partially textured mesh. Since the reprojected texture is typically incomplete due to occlusions, self-occlusions, and limited viewpoint coverage, we further apply a 3D texture completion module to infer the missing surface colors. By modeling texture directly in the 3D domain as a continuous color field, this module improves texture continuity and reduces sensitivity to UV layout and mesh topology. Together, multiview generation, texture reprojection, and 3D texture completion form a coherent pipeline that maps a single reference image to a fully textured mesh.

\subsubsection{Training Method}

The training strategy is primarily designed for the multiview generation module, with the goal of improving appearance fidelity, cross-view consistency, and robustness to illumination changes. To this end, we adopt a multiview training setup and construct training data that includes reference images, albedo supervision, and geometric conditions, while further improving data quality and diversity through data cleaning and augmentation. To better handle lighting variations commonly encountered in real inputs, we apply illumination-related augmentations during training and encourage the model to produce consistent predictions under different lighting conditions, helping it better disentangle intrinsic surface color from external illumination. The supervision combines diffusion-based training, consistency regularization, and image-level reconstruction constraints, together with a gradient-based regularization term that helps preserve local texture details and boundaries. The overall training objective is defined as
\begin{equation}
    \mathcal{L} =
    \lambda_{\text{diff}} \mathcal{L}_{\text{diff}}
    + \lambda_{\text{cons}} \mathcal{L}_{\text{cons}}
    + \lambda_{\text{img}} \mathcal{L}_{\text{img}}
    + \lambda_{\text{grad}} \mathcal{L}_{\text{grad}}.
\end{equation}
With this training strategy, the model produces stable multiview albedo predictions under varying illumination and provides reliable observations for the subsequent reprojection and texture completion stages.

\subsection{Part-Aware PBR Material Retrieval}
\label{sec:material}
We introduce a Part-Aware PBR Material Retrieval framework for generating high-quality 3D assets with PBR textures. Our pipeline couples part-based 3D segmentation with a hierarchical multi-modal retrieval strategy to ensure high physical consistency of generated materials.

The pipeline operates in three stages: First, the MatWeaver module decomposes the 3D mesh into semantically distinct components by mapping part IDs to surface faces via multi-view rendering and video-based segmentation. Second, a hierarchical retrieval engine leverages VLMs for semantic reasoning and cross-modal embedding models to identify optimal reference material samples for each part based on both visual and semantic features. Finally, a baking module extracts PBR attributes (e.g., metallic, roughness, normal maps) from these references and consolidates them into unified texture maps.

\subsubsection{Data Processing}
High-quality data preparation is the cornerstone of our Part-Aware PBR Material Retrieval framework. Our data processing pipeline is divided into two primary domains: geometric-aware segmentation for mesh understanding and a hierarchical material database for multi-modal retrieval.
\paragraph{Part Segmentation Data.}\hspace{0.25em}
To provide robust supervision for part-level segmentation, we curate a comprehensive dataset of furniture-specific 3D assets that inherently contain ground-truth part-ID annotations. We systematically transform these static meshes into synthetic video sequences, utilizing Fibonacci spherical sampling to generate consistent camera trajectories. To enhance the model's robustness, we integrate dynamic HDR environment maps and randomized lighting parameters, creating varied visual contexts. 
\paragraph{Material Database Construction.}\hspace{0.25em}
We have constructed an expert-curated database of professional interior design materials for PBR material attribute retrieval. 

The database follows a multi-level taxonomy, encompassing a wide range of primary categories—such as Fabric, Leather, and Wood, among others—which are further decomposed into granular sub-categories (e.g., Fabric is subdivided into Cotton-Linen, Lambswool, Silk, etc.). This hierarchical structure ensures that the retrieval system maintains both broad semantic relevance and fine-grained physical accuracy.

To ensure consistent and descriptive labels across this database, we employ an automated annotation pipeline. We first generate initial descriptions for each material asset using GPT-5.2, focusing strictly on intrinsic physical characteristics and textural information while intentionally ignoring color-specific attributes. However, as initial captions often suffer from low semantic distinctiveness, we further process the data by utilizing Qwen3-Embedding to cluster similar material assets within each sub-category. Finally, we leverage Gemini for contrastive rewriting: by analyzing the clustered material groups—taking both the material renderings and their initial captions as input—Gemini performs a comparative rewrite of the captions to accentuate the unique physical identifiers of each cluster. This ensures that every entry in our database is linguistically consistent, semantically unique, and highly discriminative.

\paragraph{Multi-Modal Alignment Data Synthesis.}\hspace{0.25em}
For each furniture-specific 3D mesh part, we randomly assign a material from our curated database and render it from multiple viewpoints to generate the query data. These rendered part images are then paired with detailed descriptions generated by Qwen3-VL. Simultaneously, the positive data consists of the corresponding material samples—rendered as spheres—accompanied by the refined captions derived from our automated material database pipeline.

To ensure the high fidelity and generalizability of the learned representation, we implement a rigorous quality-driven filtering pipeline. We employ a "many-to-one" mapping strategy, where each unique material ID is rendered across 50 diverse geometric shapes with multiple random viewpoints to ensure the model learns material features independent of object geometry. Crucially, we utilize Qwen3-VL for automated quality assessment to evaluate the rendering fidelity of these component images. By enforcing a confidence threshold of Score $>$ 4 based on these VLM evaluations, we curate a high-confidence subset of samples. This filtered dataset acts as the essential core for training our cross-modal embedding alignment and subsequent stage-wise model refinement.

\begin{figure}[!htbp]
    \centering
    \includegraphics[width=1.0\textwidth]{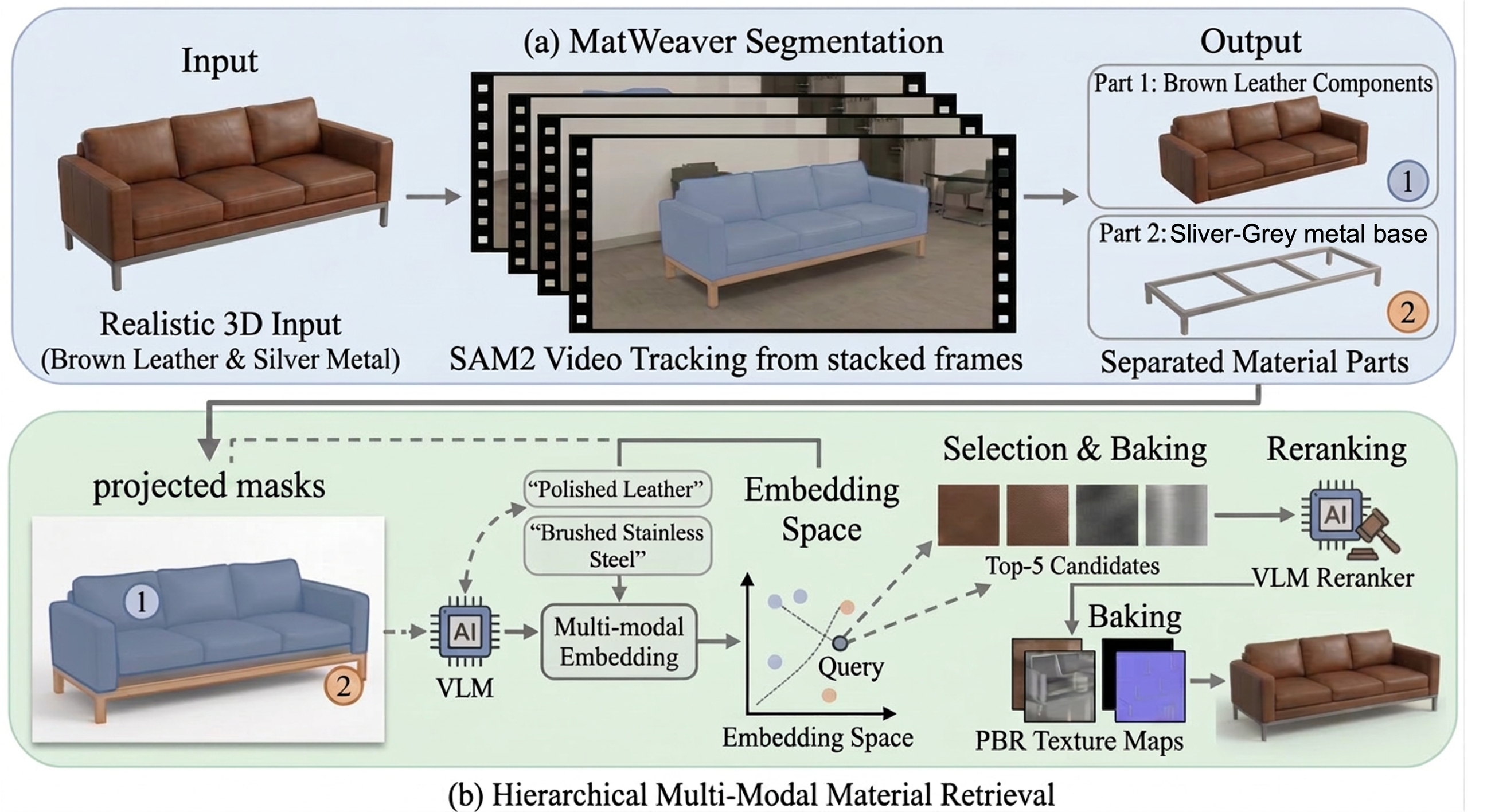} % 将此处的名称替换为你的图片文件名
    \caption{Overview of our proposed framework. The pipeline consists of two main stages: (a) \textbf{MatWeaver Segmentation}, which decomposes the 3D mesh into semantically distinct components using SAM2-based video tracking and UV-space projection; and (b) \textbf{Hierarchical Multi-Modal Material Retrieval}, which aligns 2D parts with a material library, performs cross-modal embedding search, reranks top candidates via a VLM expert judge, and finally consolidates PBR texture maps for production-ready assets.}
    \label{fig:overall_pipeline}
\end{figure}

\subsubsection{MatWeaver:Part-Aware Segmentation}

As illustrated in Fig. \ref{fig:overall_pipeline}(a), MatWeaver decomposes 3D meshes into semantically distinct components through a robust video-based segmentation framework.

\paragraph{Video-Based Semantic Segmentation.} We formulate 3D part segmentation as a video-tracking task. To initialize, we perform automated point prompt sampling on keyframes using superpixel clustering and edge-aware strategies. These prompts serve as seeds for the SAM2-based video segmentation model, which tracks and masks the target components across the entire video sequence. To ensure instance consistency, we employ an mIoU-based merging strategy: segmentation tracks with high spatial-temporal overlap are unified, and each distinct material group is assigned a globally unique ID.

\paragraph{UV-Space Projection and Baking.} To project 2D segmentation results back to 3D geometry, we construct a UV-Face Atlas—a lookup table mapping UV coordinates directly to mesh face IDs. By iterating through the video frames, we map per-pixel segmentation labels into UV space and apply a consensus voting mechanism to determine the most frequent label for each face. Finally, we perform post-processing, including hole filling and D4-alignment, to resolve discontinuities and ensure the resulting segmentation maps are topologically clean and geometrically accurate.

\subsubsection{Hierarchical Multi-Modal Material Retrieval Architecture}
Following geometric decomposition, we retrieve optimal PBR materials for each component via a hierarchical pipeline as shown in Fig. \ref{fig:overall_pipeline}(b).

\paragraph{Precision 2D-3D Alignment.} To ground our retrieval in the input studio image, we implement a pose estimation module that bridges the gap between 2D pixels and 3D geometry. We utilize RoMA for dense feature matching between the input image and synthetic renders, followed by a P4Pf solver to estimate camera parameters. We apply a three-stage optimization (coarse alignment, fine-tuning, and part-level deformation) to ensure pixel-perfect mask alignment. This allows us to project the 3D part segmentation labels directly onto the input studio image, creating localized component masks for each material slot.

\paragraph{Hierarchical Multi-Modal Retrieval.} Our retrieval engine operates through a three-stage pipeline. First, we perform Global VLM Reasoning: using the projected 2D masks, we query a VLM with the studio image to generate descriptions and material categories for each numbered component. Second, we employ Cross-Modal Embedding via Qwen3-VL: by encoding the component's visual features and its generated caption into a unified embedding space, we perform a refined retrieval within the category-specific subset of our material database, identifying the top-5 candidate material samples based on joint visual-semantic similarity. Finally, we execute VLM Reranking: we input the top-5 material candidates, the component’s original patch, and the material-specific captions into a frozen VLM. This module acts as an expert judge to resolve ambiguity and select the most physically and visually consistent material.

\paragraph{Physics-Based Baking.} Upon selecting the optimal material for each part, we consolidate the PBR attributes. We dynamically allocate independent material slots for each component to maintain logical separation. We then employ a physics-based baking workflow to combine per-component attributes ( Metallic, Roughness, Normal, etc.) into unified, engine-ready texture maps. To facilitate downstream industrial integration, we apply targeted baking and Draco compression. This process drastically reduces the memory footprint—achieving up to a 20x reduction in size—while preserving high-fidelity physical properties and ensuring the final asset is production-ready.

\subsection{Material-Aware Part Generation}
\label{sec:parts}

The parts stage decomposes the reconstructed object into semantically labelled part
geometry for independent material assignment and downstream part-level editing. Our
approach differs from pipelines that first segment an object and then generate each
part separately~\cite{xpart,seed3d2}; instead, it learns a compact part-aware latent
space and generates the complete multi-part representation in one diffusion process.
Compared with sequential per-part methods whose decoding cost grows with the number of
parts~\cite{lin2025partcrafter}, our VAE decodes multiple part-specific SDF fields in
one forward pass. Each output head corresponds to one part slot, so part identity is
carried directly by the SDF channel.

\subsubsection{Data Processing}

We collect a large set of material-aware 3D assets with part annotations and convert
all shapes into a direct multi-head SDF representation. Every part is remeshed into a
clean watertight surface before SDF supervision is generated, and the target for each
query point is stored as a multi-channel SDF vector whose channels correspond to the
part slots.

The VAE input point cloud contains positions, normals, and a part ID for each sampled
surface point. In the model, the part ID is mapped to a learned part embedding and
combined with local geometric features, making the latent code aware of both geometry
and semantic part identity. Image conditioning data is produced by rendering each
object from random viewpoints under varied environment lighting, so the DiT learns to
predict part structure from realistic single-view appearance cues. For each object, the
training record stores uniform surface samples, sharp-edge samples, and SDF queries
drawn from near-surface, near-sharp, and volumetric regions.

\subsubsection{Model Architecture}

\begin{figure}[h]
  \centering
  \includegraphics[width=0.95\linewidth]{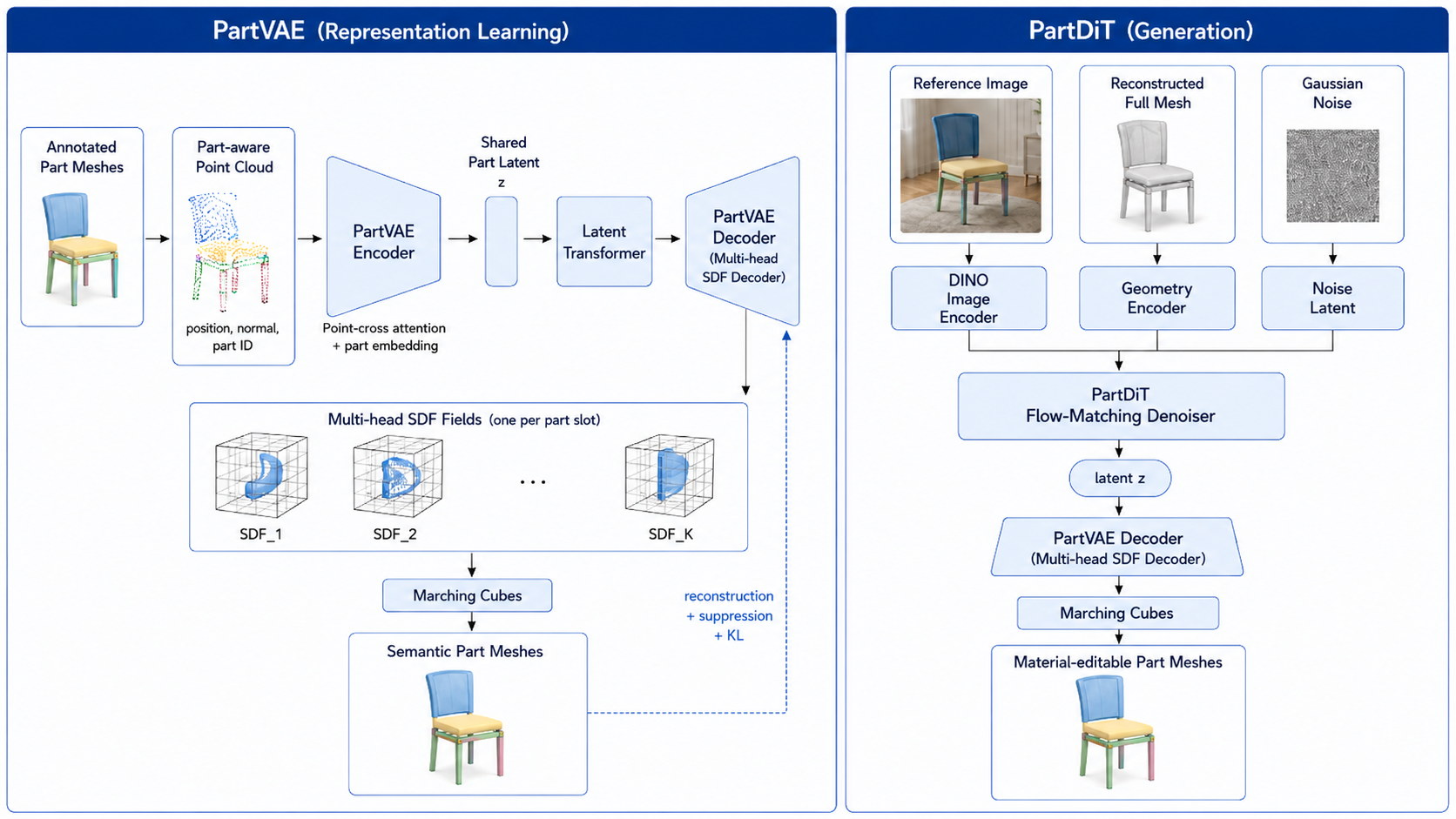}
  \caption{Architecture of the material-aware part generation stage. A PartVAE defines
  a compact latent space for multi-head part SDF fields. A PartDiT denoiser then
  generates part-aware latents conditioned on DINO image tokens and the full-object
  geometry latent.}
  \label{fig:part_architecture}
\end{figure}

The central architectural constraint of the parts module is decoding latency. A direct
part-aware extension of a 3D VAE would assign each semantic part an independent latent
slot or SDF decoder. This gives clean part separation, but it also makes representation
length, mesh extraction, and decoder execution scale with the number of parts. We
therefore design the part stage around a fixed-size multi-head SDF representation rather
than a variable-length list of per-part reconstructions.

The core mechanism for resolving this bottleneck is the \emph{PartVAE} shown on the
left side of Figure~\ref{fig:part_architecture}. The encoder follows the
point-cross-attention family of 3D VAEs~\cite{dora}, but the input point stream is made
explicitly part-aware. Each surface sample carries position and normal attributes
together with a semantic part ID. The part ID is embedded and concatenated with the
normal features before the PointCrossAttention encoder aggregates the multi-part point
cloud into one shared latent code $\mathbf{z}$ through a variational bottleneck. The
latent is shared across the whole object, but its construction is conditioned by
part-aware surface evidence.

The decoder turns this shared latent into geometry with a multi-head SDF output. A
latent transformer first refines $\mathbf{z}$, and a cross-attention decoder evaluates
the 32-head SDF channels to produce part-specific SDF fields
$\{\mathrm{SDF}_1,\ldots,\mathrm{SDF}_8\}$ in a single forward pass. An iso-surface
extraction algorithm, such as Marching Cubes, is then applied to each valid SDF channel
to recover the corresponding part mesh~\cite{marchingcubes1987}. This design keeps the
expensive implicit decoding path shared across parts: the model performs one decoder
evaluation with multiple output channels instead of launching a separate decoder for
each semantic part.

To decompose a single reconstructed object mesh into material-editable parts, we train
a DiT generator on top of the PartVAE latent space. A DINOv2-Large image encoder
extracts appearance tokens from the input image~\cite{dinov2}, while the full-object
geometry reconstruction is encoded by the VAE to provide a pure geometry condition.
Starting from Gaussian noise in the latent space, the Diffusion Transformer denoises a
complete latent code $\mathbf{z}$ conditioned on both image tokens and the geometry
latent. The generated code is decoded once by the VAE decoder into multi-head SDF
fields, yielding a set of part meshes aligned to the reference image.

\subsubsection{Training Method}

The VAE is trained with a combination of multi-head geometry reconstruction,
far-field suppression, and KL regularization:
\begin{equation}
  \mathcal{L} =
    \mathcal{L}_{\mathrm{recon}}
    + \mathcal{L}_{\mathrm{suppress}}
    + \lambda_{\mathrm{KL}}\,\mathcal{L}_{\mathrm{KL}} .
  \label{eq:part_loss}
\end{equation}
The reconstruction term supervises the per-part SDF channels on near-surface,
sharp-edge, and volumetric samples. The suppression term reduces spurious geometry in
inactive regions and encourages clean separation between part channels. The KL term
regularizes the latent space for generative modelling.

The DiT is trained with the flow matching objective with
three stages. First, a large-scale pre-training stage learns generic part generation
from the same data distribution used for geometry pre-training, giving the model broad
coverage over object categories, layouts, and part topologies. Second, a continued
training (CT) stage uses high-quality renderings with normalized lighting to strengthen
the mapping from visual material cues to semantic part structure. Finally, a supervised
fine-tuning (SFT) stage uses manually curated high-quality examples to improve boundary
cleanliness, material-part consistency, and robustness on production-facing inputs.

%% ====================================================================
\section{Evaluation}
\label{sec:evaluation}
%% ====================================================================

We evaluate Home3D 1.0 on a curated 100-case furniture benchmark. Each case contains a
reference image, a designer-authored 3D model used as ground truth, and
the generated models from Home3D 1.0 and three representative closed-source image-to-3D
systems. To ensure fair comparison across different coordinate conventions, all
generated meshes are coarsely aligned to the ground truth, refined with ICP, and then
evaluated using the same normalization, surface sampling, cameras, and lighting.

% ------------------------------------------------------------------
\subsection{Geometry Evaluation}
\label{sec:eval_geometry}
% ------------------------------------------------------------------

\begin{figure}[!htbp]
  \centering
  \includegraphics[width=\linewidth,height=0.50\textheight,keepaspectratio]{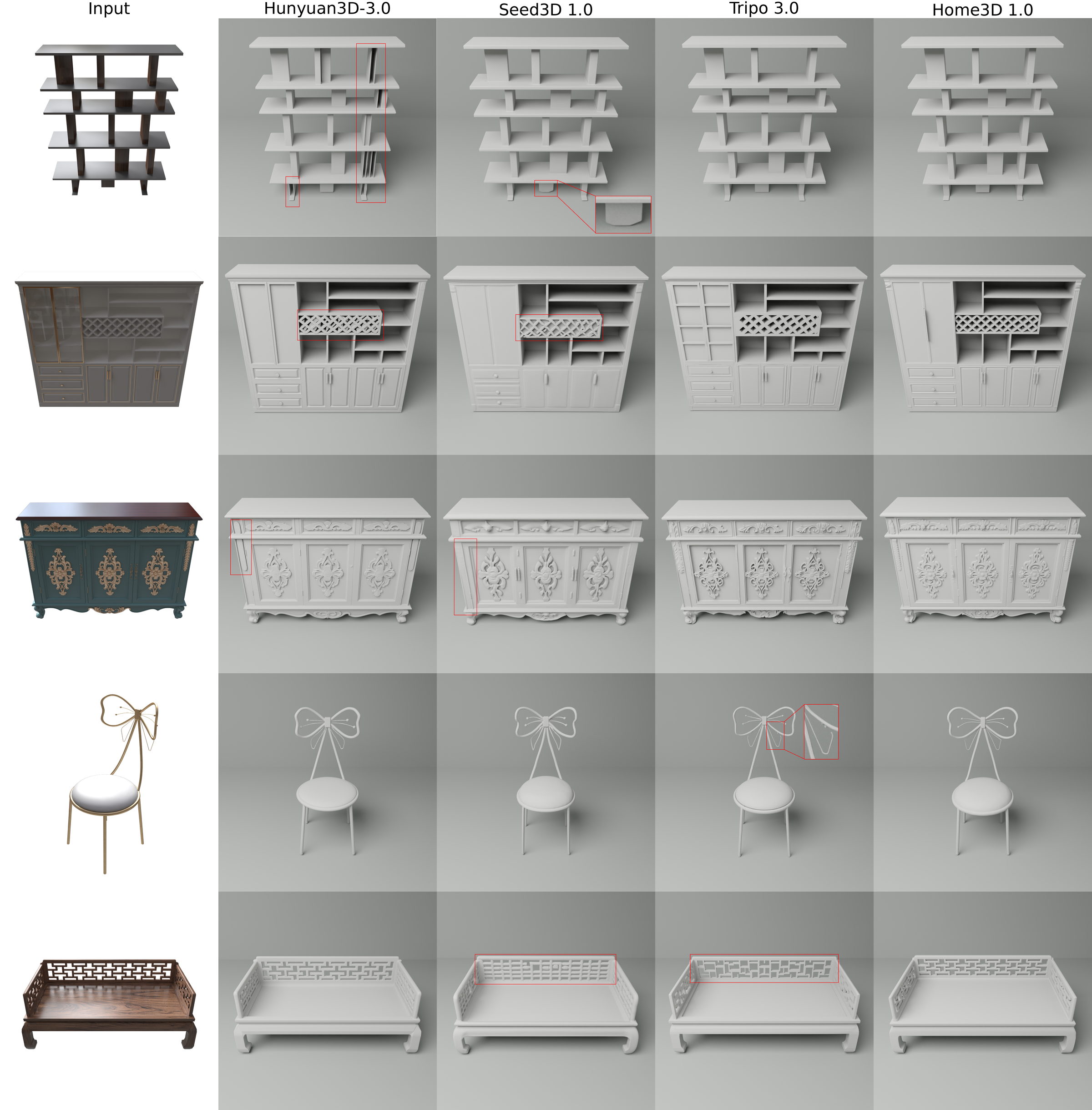}
  \caption{Qualitative comparisons between Home3D 1.0 and baselines in terms of 3D
  shape generation.}
  \label{fig:geometry_quality_comparison}
\end{figure}

For geometry evaluation, predicted and ground-truth meshes are normalized to the unit
cube $[-0.5,0.5]^3$ by bounding-box center and longest edge. We sample 30k surface
points and report Chamfer Distance (CD, $\times 10^{-3}$), Earth Mover's Distance
(EMD, $\times 10^{-2}$), and F1@0.01.
Figure~\ref{fig:geometry_quality_comparison} provides a complementary visual
comparison. While all methods recover the visible silhouette reasonably well,
Home3D 1.0 produces more complete object structure in regions that are weakly
observed or not directly visible in the input image, such as back supports, rear
legs, inner chair frames, and underside connections. This improves both global
proportions and local structural consistency, especially for furniture whose function
depends on thin supports or repeated components.
\begin{table}[H]
\centering
\small
\caption{Main evaluation results on the furniture benchmark against three closed-source image-to-3D
systems. Geometry is evaluated with CD ($\times 10^{-3}$), EMD ($\times 10^{-2}$),
and F1@0.01 after mesh alignment; appearance is evaluated from fixed-view PBR
renders with CLIP-I and LPIPS. The arrows indicate the preferred direction for each
metric.}
\label{tab:main_eval_results}
\begin{tabular}{lccccc}
\toprule
Method & CD ($\times 10^{-3}$) $\downarrow$ & EMD ($\times 10^{-2}$) $\downarrow$ & F1@0.01 $\uparrow$ & CLIP-I $\uparrow$ & LPIPS $\downarrow$ \\
\midrule
Hunyuan3D-3.0 & 0.5218 & 5.360 & 0.5987 & 0.9370 & 0.07465 \\
Seed3D 1.0    & 0.5446 & 5.215 & 0.5784 & 0.9329 & 0.08010 \\
Tripo 3.0      & 0.5646 & 5.175 & 0.5860 & 0.9324 & 0.07905 \\
\textbf{Home3D 1.0} & \textbf{0.4936} & \textbf{5.174} & \textbf{0.6329} & \textbf{0.9370} & \textbf{0.07438} \\
\bottomrule
\end{tabular}
\end{table}

Home3D 1.0 obtains the lowest CD ($\times 10^{-3}$) and EMD ($\times 10^{-2}$) among the evaluated systems, indicating the
closest surface-level geometry to designer models after normalization and alignment.
It also achieves the best F1@0.01. On rendered PBR appearance,
Home3D 1.0 leads this run in both CLIP-I and LPIPS.

% ------------------------------------------------------------------
\subsection{Texture and Material Evaluation}
\label{sec:eval_textured}
% ------------------------------------------------------------------

For texture and material evaluation, each aligned model is rendered from up to 16 fixed views
under the same HDR environment. We compute CLIP-I between corresponding rendered views
as a semantic/appearance similarity score and LPIPS~\cite{lpips2018} as a perceptual
distance. Quantitative results are reported together with the geometry metrics in
Table~\ref{tab:main_eval_results}.

\begin{figure}[!htbp]
  \centering
  \includegraphics[width=\linewidth,height=0.50\textheight,keepaspectratio]{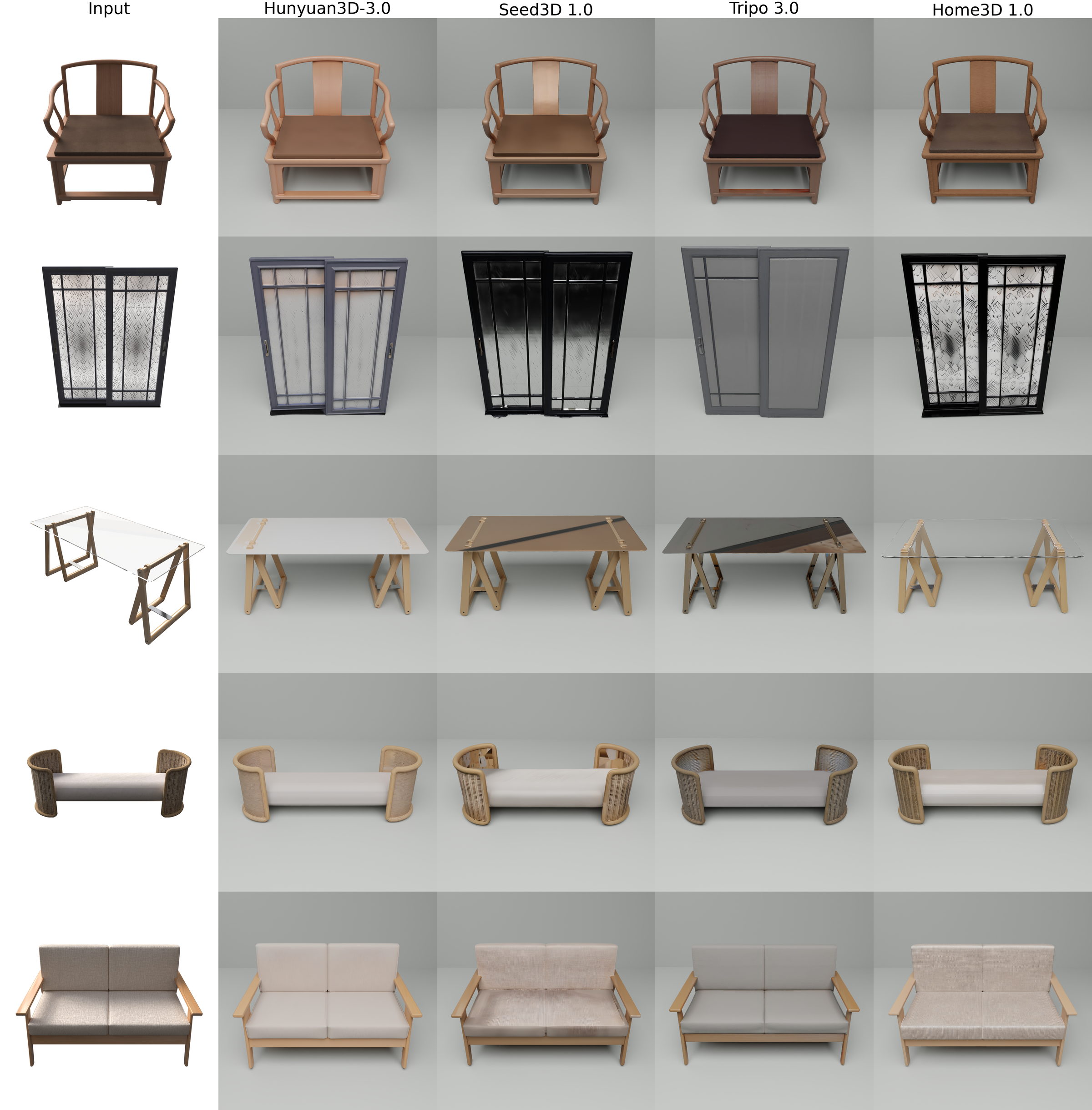}
  \caption{Qualitative comparisons between Home3D 1.0 and baselines in terms of PBR
  appearance generation.}
  \label{fig:textured_quality_comparison}
\end{figure}

The qualitative comparison shows the remaining challenges behind the rendered metrics:
material identity, fine texture detail, color consistency, and physically plausible
roughness/metallic response. Compared with the baselines, Home3D 1.0 better preserves
high-frequency surface cues such as woven fabric, wood grain, and stitched or tufted
patterns, and it more accurately represents material-specific appearance for plush and
glass furniture components. These cases are particularly sensitive to both texture
detail and PBR parameters: plush surfaces require soft, spatially varying roughness,
whereas glass requires transparent or glossy response that remains consistent across
views.

% ------------------------------------------------------------------
\subsection{Part Evaluation}
\label{sec:eval_parts}
% ------------------------------------------------------------------

For part decomposition, we evaluate the VAE independently from the image-conditioned
DiT. The VAE evaluation verifies whether the latent representation can reconstruct
multi-part geometry with fixed decoding cost while retaining semantic part support.
All methods are evaluated at $512^3$
resolution with 30k sampled points and F1 threshold 0.01.

\begin{table}[h]
\centering
\small
\setlength{\tabcolsep}{3pt}
\caption{Part-aware VAE reconstruction and decode latency}
\label{tab:part_vae_results}
\begin{tabular}{lccrcccl}
\toprule
Method &
CD ($\times 10^{4}$) $\downarrow$ &
F1@0.01 $\uparrow$ &
Avg.\ Tokens &
\multicolumn{3}{c}{Enc-Dec Time (s) $\downarrow$} &
Part Support \\
\cmidrule(lr){5-7}
 & & & & 1 part & 16 parts & 32 parts & \\
\midrule
Hunyuan3D 2.1\cite{hunyuan3d21} & 2.589 & 68.683 & 4,096 & 14.387 & -- & -- & none \\
TRELLIS.2\cite{xiang2025trellis2}     & 1.844 & 69.609 & 2,239 & 0.861 & -- & -- & none \\
PartCrafter\cite{lin2025partcrafter}     & 9.946 & 55.499 & 52,880 & 6.807 & 48.968 & 101.529 & semantic \\
PartPacker\cite{partpacked}    & 3.529 & 70.304 & 8,192 & 30.231 & 31.822 & 33.993 & spatial only \\
\midrule
\textbf{PartVAE} & 3.478 & 66.094 & 8,192 & 16.344 & 21.074 & 24.048 & semantic \\
\bottomrule
\end{tabular}
\end{table}

\begin{figure}[!htbp]
  \centering
  \includegraphics[width=0.92\linewidth]{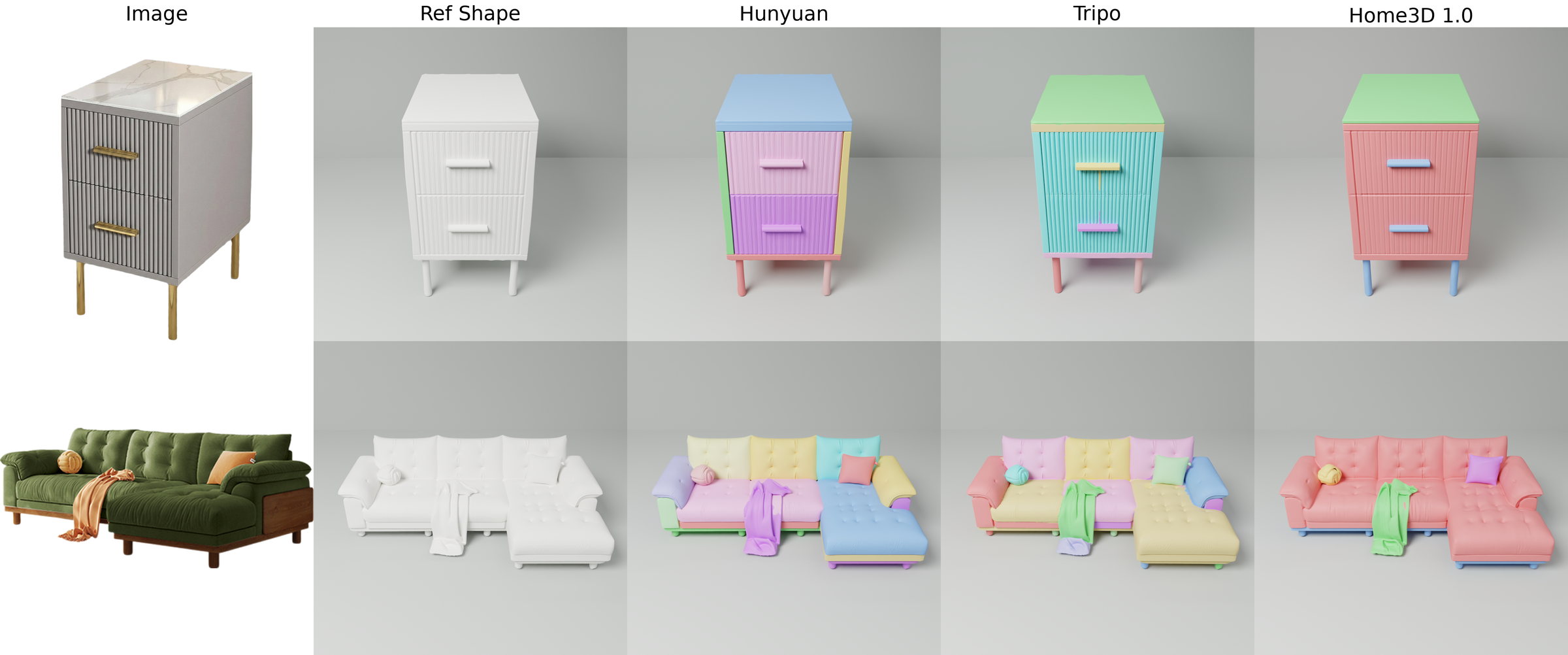}
  \caption{Qualitative comparisons of material-aware part decomposition against
  closed-source image-to-3D systems. Home3D 1.0 separates furniture into
  material-consistent editable parts, while the compared systems mainly recover a
  single textured surface or entangle different materials within the same region.}
  \label{fig:material_aware_part_methods}
\end{figure}

\begin{figure}[!htbp]
  \centering
  \includegraphics[width=0.92\linewidth]{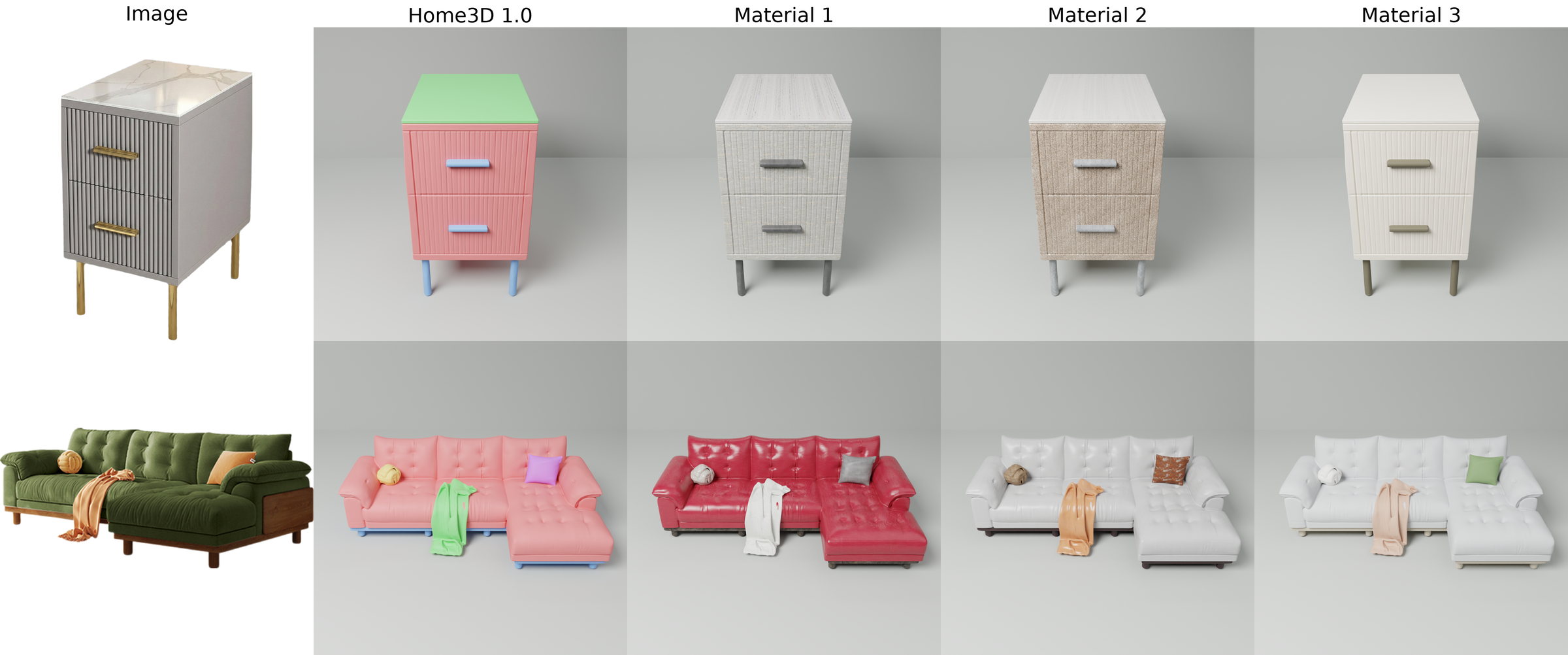}
  \caption{Material replacement using Home3D 1.0 part decomposition. After
  material-aware part generation, individual components can be assigned different
  fabric, wood, metal, or plastic appearances.}
  \label{fig:material_aware_part_editing}
\end{figure}

The Part-Aware VAE is not optimized only for lowest holistic CD. Its practical value is
that part SDF fields are decoded directly from one shared latent: unlike sequential
part generators, Home3D 1.0 does not generate parts through repeated per-part decoder
calls. The qualitative comparison in
Figure~\ref{fig:material_aware_part_methods} shows that Home3D 1.0 produces cleaner
material-aware part splits than closed-source image-to-3D systems. As shown in
Figure~\ref{fig:material_aware_part_editing}, this decomposition is directly useful
in home-design workflows: by splitting furniture into material-editable regions with
semantic labels, the generated asset allows designers to quickly replace fabrics,
woods, metals, glass, or other component-level materials without manually
re-segmenting the whole model.

%% ====================================================================
\section{Conclusion}
\label{sec:conclusion}
%% ====================================================================

We have presented Home3D, a four-stage image-to-3D pipeline for interior design.
The pipeline covers geometry reconstruction, PBR texture synthesis, material segmentation
and retrieval, and per-part mesh decomposition, with each stage evaluated independently.
Key contributions include a 3D implicit texture function that sidesteps UV instability,
a video-based UV-voting segmentation requiring no manual annotation, and a multi-head
SDF decoder that reconstructs semantic part fields in one forward pass.
Future work will focus on improving robustness for broader deployment, including
cross-stage feedback---e.g., using material predictions to guide geometry decimation---
and extending the parts module to articulated and deformable objects.

\bibliographystyle{plain}
\bibliography{create3d}

\clearpage
\appendix

%% ====================================================================
\section{Contributions and Acknowledgments}
\label{app:contributions}
%% ====================================================================

All contributors of Home3D 1.0 are listed in alphabetical order by their last names.

\subsection{Core Contributors}

Yiyun Fei, Guoqiu Li, Jin Song, Chuqiao Wu, Delong Wu, Hong Wu, Ziru Zeng

\subsection{Contributors}

Haohui Chen, Yindong Kong, Jing Li, Qi Wu, Feng Zhang

\subsection{Acknowledgments}

Jianan Jiang, Shichen Lv, Kui Wang, Ruigao Yang

\end{document}